\documentclass[10pt,twocolumn,letterpaper]{article}

\usepackage{iccv}
\usepackage{times}
\usepackage{epsfig}
\usepackage{graphicx}
\usepackage{amsmath}
\usepackage{amssymb}

\usepackage{xcolor}
\usepackage{colortbl}
\usepackage{float}
\usepackage{multirow}
\usepackage{theorem}

\usepackage{algorithmic}
\usepackage{algorithm}
\usepackage{pifont}
\newcommand{\cmark}{\ding{51}}%

\definecolor{LightYellow}{RGB}{255,242,204}
\definecolor{LightGray}{gray}{0.93}
\newcolumntype{a}{>{\columncolor{LightGray}}c}

\newcommand\Tstrut{\rule{0pt}{2ex}}         
\newcommand\Bstrut{\rule[-1ex]{0pt}{0pt}}   
\newcommand\Tstrutlarge{\rule{0pt}{3.5ex}}         
\newcommand\Bstrutlarge{\rule[-2.5ex]{0pt}{0pt}}   

\usepackage[pagebackref=true,breaklinks=true,colorlinks,bookmarks=false]{hyperref}

\iccvfinalcopy 

\def\iccvPaperID{9709} 

\ificcvfinal\fi

\begin{document}

\hbadness=2000000000
\vbadness=2000000000
\hfuzz=100pt

\setlength{\abovedisplayskip}{1pt}
\setlength{\belowdisplayskip}{1pt}
\setlength{\floatsep}{3pt plus 1.0pt minus 1.0pt}
\setlength{\intextsep}{3pt plus 1.0pt minus 1.0pt}
\setlength{\textfloatsep}{3pt plus 1.0pt minus 1.0pt}
\global\setlength{\theorempreskipamount}{3pt plus 1.0pt minus 1.0pt}
\global\setlength{\theorempostskipamount}{3pt plus 1.0pt minus 1.0pt}
\setlength{\parskip}{0pt}

\title{Augmenting and Aligning Snippets for Few-Shot Video Domain Adaptation}

\author{Yuecong Xu, Jianfei Yang, Yunjiao Zhou\\
School of Electrical and Electronic Engineering, Nanyang Technological University, Singapore\\
50 Nanyang Avenue, Singapore 639798\\
{\tt\small \{xuyu0014, yang0478, yunjiao001\}@e.ntu.edu.sg}
\and
Zhenghua Chen, Min Wu, Xiaoli Li\\
Institute for Infocomm Research (I\textsuperscript{2}R), A*STAR, Singapore\\
1 Fusionopolis Way, \#21-01, Connexis South, Singapore 138632\\
{\tt\small \{chen\textunderscore zhenghua, wumin\}@i2r.a-star.edu.sg}
}

\maketitle
\ificcvfinal\thispagestyle{empty}\fi

\begin{abstract}
   For video models to be transferred and applied seamlessly across video tasks in varied environments, Video Unsupervised Domain Adaptation (VUDA) has been introduced to improve the robustness and transferability of video models. However, current VUDA methods rely on a vast amount of high-quality unlabeled target data, which may not be available in real-world cases. We thus consider a more realistic \textit{Few-Shot Video-based Domain Adaptation} (FSVDA) scenario where we adapt video models with only a few target video samples. While a few methods have touched upon Few-Shot Domain Adaptation (FSDA) in images and in FSVDA, they rely primarily on spatial augmentation for target domain expansion with alignment performed statistically at the instance level. However, videos contain more knowledge in terms of rich temporal and semantic information, which should be fully considered while augmenting target domains and performing alignment in FSVDA. We propose a novel SSA\textsuperscript{2}lign to address FSVDA at the snippet level, where the target domain is expanded through a simple snippet-level augmentation followed by the attentive alignment of snippets both semantically and statistically, where semantic alignment of snippets is conducted through multiple perspectives. Empirical results demonstrate state-of-the-art performance of SSA\textsuperscript{2}lign across multiple cross-domain action recognition benchmarks.
\end{abstract}

\section{Introduction}
\label{section:intro}

Video Unsupervised Domain Adaptation (VUDA)~\cite{chen2019temporal,choi2020shuffle,xu2022aligning,wang2022calibrating,xu2022source} aims to improve the generalizability and robustness of video models by transferring knowledge to new domains, and is widely applied in scenarios where massive labeled videos are unavailable. Current VUDA methods assume that sufficient target data are accessible which enables domain alignment by minimizing cross-domain distribution discrepancies and obtaining domain invariant representations~\cite{chen2019temporal,choi2020shuffle,xu2023multi}. However, this assumption may not be feasible in real-world applications such as in smart hospitals and security surveillance where video models are leveraged for anomaly behavior recognition~\cite{sultani2018real,said2021efficient}, and are expected to be functional at all times even across different environments. It is more practical to obtain a few labeled videos during the early stage of model deployment to improve the transferred models' performances in the new (target) environment. A \textit{Few-Shot Video Domain Adaptation} (\textit{FSVDA}) task is hence formulated to enable knowledge learned from labeled source video to be transferred to the target video domain given only very limited labeled target videos.

With only several target domain samples, FSVDA is much more challenging than VUDA, since aligning distributions with limited samples is harder. A few research has touched on the image-based Few-Shot Domain Adaptation (FSDA)~\cite{motiian2017few,wang2019few,xu2019dsne,gao2022acrofod} by domain alignment, e.g., moment matching or adversarial training, between a spatial-augmented target domain and a filtered target-similar source domain. More recently, there have been a few early research on FSVDA~\cite{gao2020pairwise-tip,gao2020pairwise-tnnls} which extends the above strategies to videos by viewing each video sample as a whole and obtaining frame-based video features.

However, there are two major shortcomings when the image-based FSDA is applied to video domains. Firstly, applying frame-level spatial augmentation towards individual video frames ignores and undermines temporal correlation across sequential frames, and we find that such augmentation would result in only minor or even negative effects on FSVDA performance. Secondly, the effectiveness of domain alignment methods is built upon sufficient source domain and target domain data that depicts the distribution discrepancy, which is not available in FSVDA. Even worse, statistical estimation of video data distribution is less accurate due to the complicated temporal structure of video data. In this paper, we aim to overcome these two challenges by designing more effective target domain augmentation and semantic alignment in the spatial-temporal domain.

To this end, we propose to address the FSVDA task by a \textbf{S}nippet-attentive \textbf{S}emantic-statistical \textbf{Align}ment with \textbf{S}tochastic \textbf{S}ampling \textbf{A}ugmentation (\textbf{SSA\textsuperscript{2}lign}). Instead of aligning features of whole video samples at the video level or frame level~\cite{gao2020pairwise-tip,gao2020pairwise-tnnls}, we align source and target video features at the snippet level. Snippets are formed from a limited series of adjacent sequential frames, thus they contain both spatial and short-term temporal information. Leveraging snippet features for FSVDA brings two unique advantages: i) a larger amount of target domain samples could be obtained via spatial-temporal augmentations on snippets, obtaining more diverse features across the temporal dimension; ii) additional alignment of the diverse but highly correlated snippet features of each video could further improve the discriminability of the corresponding videos, which has been proven to benefit the effectiveness of video domain adaptation~\cite{chen2019transferability,yang2020mind,kundu2022balancing,xu2022source}. SSA\textsuperscript{2}lign is therefore proposed. It firstly augments the source and target domain data by a simple yet effective stochastic sampling process that makes full use of the abundance of snippet information and then performs semantic alignment from three perspectives: alignment based on semantic information within each snippet, cross-snippets of each video, and across snippet-level data distribution. Our method is demonstrated to be very effective for the FSVDA problem, outperforming the state-of-the-art methods by a large margin on two large-scale VUDA benchmarks.

In summary, our contributions are threefold. (i) We propose a novel SSA\textsuperscript{2}lign to address FSVDA at the snippet level by both statistical and semantic alignments that are achieved from three perspectives. (ii) We propose to augment target domain data and the snippet-level alignments by a simple yet effective stochastic sampling of snippets for more robust video domain alignment. (iii) Extensive experiments show the efficacy of SSA\textsuperscript{2}lign, achieving a remarkable average improvement of $13.1\%$ and $4.2\%$ over current state-of-the-art FSDA/FSVDA methods on two large-scale cross-domain action recognition benchmarks.

\section{Related Work}
\label{section:related}

\noindent \textbf{(Video) Unsupervised Domain Adaptation ((V)UDA).}
Current UDA and VUDA methods aim to transfer knowledge from the source to the target domain given that both domains contain sufficient data, improving the transferability and robustness of models~\cite{xu2022videosurvey,yang2022deepsurvey}. They could be generally divided into four categories: a) reconstruction-based methods~\cite{ghifary2016deep,yang2020label}, where domain-invariant features are obtained by encoders trained with data-reconstruction objectives; b) adversarial-based methods~\cite{chen2019temporal,xu2022aligning,choi2020shuffle}, where feature generators obtain domain-invariant features leveraging domain discriminators, trained jointly in an adversarial manner~\cite{huang2011adversarial,ganin2015unsupervised}; c) semantic-based methods~\cite{yang2021advancing,xu2022source}, which exploit the shared semantics across domains such that domain-invariant features are obtained; and d) discrepancy-based methods \cite{saito2018maximum,zhang2019bridging}, which mitigate domain shifts by applying metric learning, minimizing metrics such as MMD~\cite{long2015learning} and CORAL~\cite{sun2016return}. With the introduction of cross-domain video datasets such as Daily-DA~\cite{xu2023multi} and Sports-DA~\cite{xu2023multi}, there has been a significant increase in research interest for VUDA~\cite{choi2020shuffle,pan2020adversarial,chen2020action}. Despite the gain in video model robustness thanks to VUDA methods, they all assume that sufficient target data are accessible, which may not be feasible in real-world cases where a large amount of superior unlabeled target data are not available.

\noindent \textbf{Few-Shot (Video) Domain Adaptation (FS(V)DA).}
It is more practical to obtain a few labeled target data to aid video models to adapt. There have been a few research that explores image-based FSDA. Among them, FADA~\cite{motiian2017few} is adversarial-based and augments the domain discriminator to classify 4 types of source-target pairs. d-SNE~\cite{xu2019dsne} learns a latent space through SNE~\cite{hinton2002stochastic} with large-margin nearest neighborhood~\cite{domeniconi2005large}, and utilizes spatial augmentations to create sibling target samples. AcroFOD~\cite{gao2022acrofod} explores FSDA for object detection by applying multi-level spatial augmentation and filtering target-irrelevant source data. There are also works as in \cite{zhao2021domain,teshima2020few,sun2021anomaly,yue2021multi} that combine domain adaptation (DA) with few-shot learning (FSL), yet we differ them in the assumption of similar target and source classes and only limited target data accessible, which is more realistic. More recently, there have been a few early research on FSVDA, including PASTN~\cite{gao2020pairwise-tip} that constructs pairwise adversarial networks performed across source-target video pairs, while PTC~\cite{gao2020pairwise-tnnls} further leverages optical flow features. Both PASTN and PTC obtain video features from a frame-based video model. Despite some advances made in FS(V)DA, the above methods have not tackled FSVDA effectively by leveraging the rich temporal information as well as semantic information embedded within videos. We propose to engage in FSVDA by augmenting and attentively aligning snippet-level features which contain temporal information via both semantic and statistical alignments.

\section{Proposed Method}
\label{section:method}
For \textit{Few-Shot Video Domain Adaptation}, we are given a labeled source domain $\mathcal{D}_{S}=\{(V_{S, i},y_{S, i})\}^{N_S}_{i=1}$ with sufficient $N_{S}$ i.i.d.\ source videos across $\mathcal{C}$ classes, characterized by a probability distribution of $p_{S}$. We are also given a labeled target domain $\mathcal{D}_{T}=\{(V_{T, j},y_{T, j})\}^{N_T}_{j=1}$ with a limited number of $N_{T}\ll N_{S}$ i.i.d.\ target videos across the same $\mathcal{C}$ classes, where each video class only contains $k$ target video samples (corresponding to the $k$-shot Video Domain Adaptation task), thus $N_{T}=k\times \mathcal{C}$. $\mathcal{D}_{T}$ is characterized by a probability distribution of $p_{T}$.

Owing to the absence of sufficient target data and the lack of target information, FSVDA is much more challenging than VUDA. Current VUDA methods~\cite{chen2019temporal,xu2022aligning} that are primarily moment matching-based are ineffective without target information for domain alignment. FSVDA should be tackled by leveraging the temporal information of videos fully for more temporally diverse features while aligning with the embedded semantic information to improve video discriminability for effective video domain adaptation. We propose SSA\textsuperscript{2}lign, a novel method to transfer knowledge from the source domain to the target domain with only limited labeled target data by obtaining, augmenting, and aligning snippet features attentively. We start by introducing how snippet features are obtained and augmented through the Stochastic Sampling Augmentation (SSA), followed by a detailed illustration of the proposed SSA\textsuperscript{2}lign.

\subsection{Snippet Features with the Stochastic Sampling Augmentation}
\label{section:method:snippet-ssa}

\begin{figure*}[!ht]
\begin{center}
   \includegraphics[width=1.\linewidth]{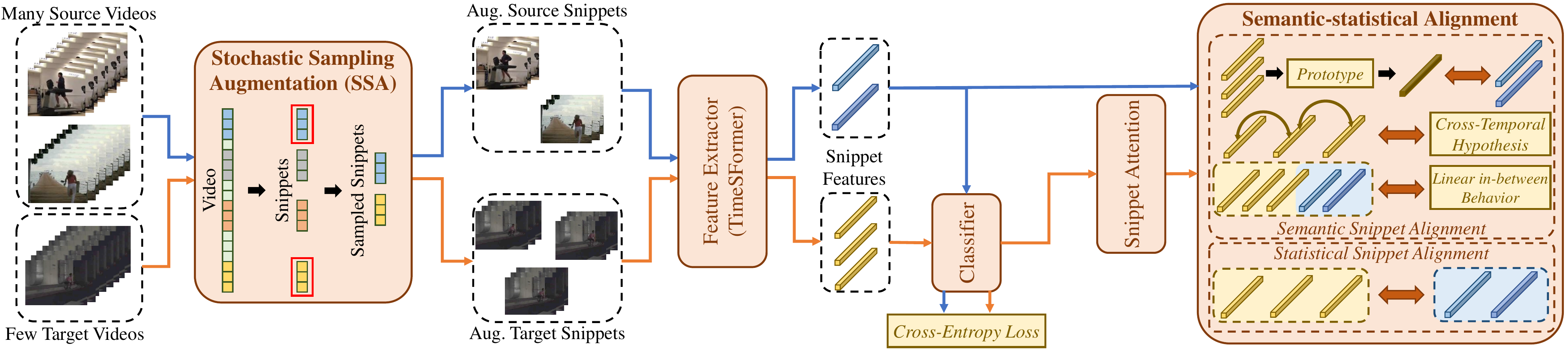}
\end{center}
\vspace{-0.5em}
\caption{Pipeline of SSA\textsuperscript{2}lign. Source and target snippets are first obtained through the Stochastic Sampling Augmentation, whose features are obtained through the shared feature extractor. SSA\textsuperscript{2}lign then aligns the source and target domains at the snippet level with the Semantic-statistical Alignment, while weighing the impact of different target snippets through snippet attention, whose weight is built based on the output prediction of target snippets, obtained from a shared classifier with source snippets. The blue and orange lines imply the data flow for source and target videos respectively.
}
\vspace{-1em}
\label{figure:3-1-struct}
\end{figure*}

The key to effective target domain expansion and domain alignment in FSVDA is to obtain and augment features with temporal information such that the augmented features are diverse temporally. While various spatial augmentation methods (e.g., color jittering, flipping, cropping) have been adopted in supervised action recognition thanks to their capability in improving the robustness of video models, and in prior FSDA for expanding the target domain $\mathcal{D}_{T}$, they are performed at the frame-level across randomly selected individual frames. Meanwhile, the temporal information corresponds to the correlation of sequential frames and would be undermined by spatial augmentation since sequential frames may not be equally augmented. Augmentations for FSVDA must be performed above the frame level.

Snippets are formed from a limited series of adjacent sequential frames and have been utilized in multiple supervised action recognition methods (e.g., TSN~\cite{wang2018temporal} and STPN~\cite{wang2017spatiotemporal}) thanks to their ability in including both spatial and short-term temporal information. Therefore, we align source and target video features at the snippet level. Mathematically, given a target video $V = [f^{1}, f^{2}, ..., f^{n}]$ that contains $n$ frames, we denote the $i$-th frame as $f^{i}$. We denote the length of a snippet $s$ to be $m$, then video $V$ would contain $n-m+1$ snippets in total. We define a snippet $s^{j} = [f^{j}, f^{j+1}, ..., f^{j+m-1}]$ as the snippet starting from the $j$-th frame. While given only $N_{T}=k\times \mathcal{C}$ target videos, there are $N_{T}\times (n-m+1)$ target snippets, which can greatly expands the target domain for domain alignment while preserving essential temporal information.

While the target domain is largely expanded, utilizing all snippets for alignment is computationally inefficient (a 10-second 30-fps video contains more than 290 8-frame snippets). Moreover, snippets that are obtained adjacently would differ over only ONE frame, resulting in high redundancy in temporal information. To ensure that diverse temporal information is utilized, we adopt a simple Stochastic Sampling Augmentation (SSA) over the snippets. Formally, during training we sample $r>1$ snippets $s^{a}_{b}$ stochastically per target video per mini-batch, where $a\in[1, n-m+1]$ denotes the starting frame of the snippet and $b\in[1, r]$ denotes the $b$-th snippet sampled. SSA further ensures that the sampled snippets are diverse from two perspectives. Firstly, SSA samples snippets with a minimum of $\hat{m}$ difference between the starting frame of any two snippets from the same target video, that is $\forall b_{x}\in[1, r], b_{y}\in[1, r]$ with $b_{x}\neq b_{y}$, we set $|a_{x}-a_{y}|\geqslant \hat{m}$. Secondly, since there are much more source videos than target videos during training, it is likely that the same target video would be encountered across different mini-batches. SSA ensures that different snippets are sampled each time the same target video is included in a mini-batch across the same training epoch.

The SSA is also applied to the source videos to obtain source snippets. However, since there are sufficient source videos, it is more reasonable and efficient to exploit source knowledge with different source videos rather than the different snippets of a source video that would contain redundant source knowledge. Therefore, we only sample $r=1$ snippet stochastically per source video via SSA.

Another crucial step towards transferring source knowledge to the target domain is to obtain rigorous snippet features that include both spatial and temporal information. We resort to the Transformer-based TimeSFormer~\cite{bertasius2021space} which extracts spatial and temporal features with separate space-time attention blocks based on self-attention~\cite{vaswani2017attention}. While various Transformer-based video models achieve competitive performances on action recognition, TimeSFormer possesses the least amount of parameters, requiring only $60\%$ parameters of Swin~\cite{liu2022video} and only $40\%$ parameters of ViViT~\cite{arnab2021vivit}. The feature of snippet $s^{a}_{b}$ is $\mathbf{f}_{b} = Time(s^{a}_{b})$ where $Time$ denotes the TimeSFormer. 

\subsection{Snippet-attentive Semantic-statistical Alignment with SSA}
\label{section:method:snippet-ssa2}
With the absence of sufficient target data, conventional VUDA methods that are primarily moment matching-based would not be fully effective since target data distribution is unknown. Alternatively, we tackle FSVDA at the snippet level by aligning the embedded semantic information from three perspectives: aligning based on the semantic information within each snippet, cross-snippets of each video, and across snippet-level data distribution. Statistical alignment is also adopted for more stable domain alignment, while both alignments attend to the more impactful snippets. 

Following the above strategy, we propose the \textbf{Snippet-attentive Semantic-statistical Alignment (SSAlign)}, with the input obtained through SSA introduced in Sec. \ref{section:method:snippet-ssa}, forming the \textbf{SSA\textsuperscript{2}lign}. The overall pipeline of SSA\textsuperscript{2}lign is presented in Fig. \ref{figure:3-1-struct}. We obtain the augmented source and target snippets through SSA whose features are extracted by applying TimeSFormer. We denote a source snippet from the $i$-th source video as $s_{S, i}$ and its feature as $\mathbf{f}_{S, i}$, while the $l$-th target snippet ($l\in[1, r]$) from the $j$-th target video as $s_{T, jl}$ and its feature as $\mathbf{f}_{T, jl}$. The superscript of the snippet expression is omitted for clarity. Domain alignment is achieved by performing both the Semantic Snippet Alignment and the Statistical Snippet Alignment. The snippet attention is applied to the augmented target snippets to weigh the snippets dynamically. The TimeSFormer feature extractor $Time$ is shared across source and target domains while a shared classifier $H$ outputs a prediction $o$ for the source and target snippets, optimized through a cross-entropy loss:
\begin{equation}
\label{eqn:method:ce-loss-all}
\resizebox{.75\linewidth}{!}{$
\begin{aligned}
\mathcal{L}_{pred} &= \frac{1}{N_{S}} \sum\nolimits_{i=1}^{N_{S}} l_{ce}({o}_{S, i}, y_{S, i}) \\
&+ \frac{1}{N_{T}\times r} \sum\nolimits_{j=1}^{N_{T}} \sum\nolimits_{l=1}^{r} l_{ce}({o}_{T, jl}, y_{T, j}),
\end{aligned}
$}
\end{equation}
where ${o}_{S, i} = \sigma(H(\mathbf{f}_{S, i}))$ and ${o}_{T, jl} = \sigma(H(\mathbf{f}_{T, jl}))$ are the output predictions of snippet features $\mathbf{f}_{S, i}$ and $\mathbf{f}_{T, jl}$, while $\sigma$ denotes the SoftMax function.

\noindent \textbf{Semantic Snippet Alignment.}
The purpose of applying semantic alignment at the snippet level is to match the embedded semantic information (e.g.,\ each individual feature or characteristic over a set of features) across source and target domains. Since both domains share the same TimeSFormer feature extractor, this implies that for each individual snippet feature, those of the same class should be close together across both domains. However, it is computationally expensive to compute the distances between each source and target snippet features given their large quantity. Inspired by the Prototypical Network~\cite{snell2017prototypical,liu2020prototype} designed for few-shot learning~\cite{zhu2021few,wang2019panet}, we resort to a more efficient solution where semantic alignment across each snippet is performed by minimizing the distance between source snippet features and target prototypes. The target prototypes are obtained for each individual class $C_{x}$ as the mean feature of all target snippet features classified as $C_{x}$, formulated as:
\begin{equation}
\label{eqn:method:proto-tgt}
Pr_{x} = \frac{1}{n_{T, x}} \sum\nolimits_{\forall s_{T, jl}\in C_{x}} \mathbf{f}_{T, jl},
\end{equation}
where $n_{T, x}$ is the number of target snippets classified as class $C_{x}$. For stable and effective alignment, the snippet features for computing the target prototypes are obtained after $e$ training epochs. Target prototypes are subsequently updated per epoch by their exponential moving average as:
\begin{equation}
\label{eqn:method:proto-update-tgt}
Pr_{x} \gets \lambda_{P}Pr_{x} + (1-\lambda_{P})Pr'_{x},
\end{equation}
where $Pr_{x}$ and $Pr'_{x}$ denote the target prototype of class $C_{x}$ computed at the current and previous epochs. Aligning source snippet features towards target prototypes is thus achieved by minimizing the Euclidean distances between them and denoted as the prototype alignment loss as:
\begin{equation}
\label{eqn:method:proto-align-loss}
\mathcal{L}_{proto} = \frac{1}{N_{S}} \sum\nolimits_{x=1}^{\mathcal{C}} \sum\nolimits_{i=1}^{n_{S, x}} \sqrt{(\mathbf{f}_{S, i} - Pr_{x})^{2}}.
\end{equation}
$n_{S, x}$ is the number of source snippets classified as class $C_{x}$. 

Besides the capability of obtaining temporally diverse features via SSA, leveraging snippet features for FSVDA is also more advantageous due to the inclusion of additional semantic information that exists across the diverse but highly correlated snippet features obtained from the same video, which should also be aligned. However, since we aim to exploit more source information with different source videos, the source cross-snippet semantic information cannot be directly obtained. Alternatively, the \textit{cross-temporal hypothesis} introduced in~\cite{xu2022source} provides a thorough description of the cross-snippet semantic information for the source videos.
Therefore, the equivalence of aligning the cross-snippet semantic information across source and target domains is to align the cross-snippet semantic information of the target domain to the \textit{cross-temporal hypothesis}, that is the snippet features over the snippets obtained from the same target video through SSA must be consistent. Meanwhile, aligning the \textit{cross-temporal hypothesis} would also drive target videos to be discriminative, while previous studies~\cite{chen2019transferability,yang2020mind,kundu2022balancing,xu2022source} have proven that improving discriminability can benefit the effectiveness of domain adaptation.

Formally, the cross-snippet consistency is achieved by minimizing the Kullback–Leibler (KL) divergence of the predictions of target snippets corresponding to the same target video. It is computed between each snippet against the key snippet of each target video, which is identified such that it is classified correctly and is certain in its prediction (i.e., low prediction entropy). In cases where no snippets are classified correctly, the snippet with the lowest prediction entropy is identified as the key snippet. The cross-snippet consistency loss is computed as:
\begin{equation}
\label{eqn:method:const-cross-loss}
\resizebox{.9\linewidth}{!}{
$\mathcal{L}_{cross} = \frac{1}{N_{T} (r-1)} \sum\limits_{j=1}^{N_{T}}  \sum\limits_{l=1, l\neq k}^{r} KL(\log({o}_{T,jy}) \| \log({o}_{T, jl}))$,
}
\end{equation}
where $KL(p\|q)$ denotes the KL-divergence while $y$ denotes $y$-th snippet corresponding to the target video $V_{T,j}$ identified as the key snippet.

Aligning semantically via matching the characteristics over differed snippet features could be further performed across the snippet-level data distribution. Since source snippets for training are obtained stochastically at each training epoch, semantic information embedded across the source snippet-level data distribution changes continuously, and would therefore be ineffective for the target snippet-level data distribution to be directly aligned. Alternatively, snippet features that are highly discriminative would imply effective domain adaptation since it has been proven that improving discriminability benefits domain adaptation~\cite{chen2019transferability,yang2020mind,kundu2022balancing,xu2022source}. We thus aim to drive the feature extractor towards obtaining snippet features that are distributed more discriminatively. Specifically, results in model robustness~\cite{zhang2018mixup} suggest that the discriminability of features can be improved if the feature extractor behaves linearly in-between training samples. The linear in-between behavior can be complied by employing the interpolation consistency training (ICT) technique~\cite{verma2019interpolation} across both source and target snippets, which encourages the linearly interpolated features to produce a linearly interpolated prediction. Formally, given a pair of snippet features $\mathbf{f}_{\ast}$, $\mathbf{f}_{\ast'}$, and their corresponding output predictions ${o}_{\ast}$, ${o}_{\ast'}$, the ICT is conducted with the following process and optimization loss:
\begin{equation}
\label{eqn:method:ict-tgt-one}
\begin{aligned}
&\mathbf{\Tilde{f}} = \lambda_{v} \mathbf{f}_{\ast} + (1 - \lambda_{v}) \mathbf{f}_{\ast'}.\\
&\mathbf{\Tilde{o}} = \lambda_{v} \mathbf{o}_{\ast} + (1 - \lambda_{v}) \mathbf{o}_{\ast'}.\\
&\mathcal{L}_{ICT} (\ast, \ast') = l_{ce}(\sigma(H(\mathbf{\Tilde{f}})), \mathbf{\Tilde{o}}),
\end{aligned}
\end{equation}
where $\lambda_{v} \in Beta(\alpha_{v}, \alpha_{v})$ is the weight assigned to $\mathbf{f}_{T, j_{1}l_{1}}$ sampled from a Beta distribution with $\alpha_{v}$ as the parameter. We refer to previous works~\cite{liang2022dine,xu2022extern} and set $\alpha_{v}=0.3$. $\mathbf{\Tilde{f}}$ and $\mathbf{\Tilde{o}}$ are the linearly interpolated features and the interpolated output predictions. In practice, we drive snippets to comply with the linear in-between behaviour by forming a single stochastic snippet pair for every snippet, forming $(N_{T}\times r+N_{S})$ snippet pairs. Aligning the snippet-level data distribution with the linear in-between behavior is achieved by optimizing the snippet distribution loss as:
\begin{equation}
\label{eqn:method:snippet-dist-tgt}
\resizebox{.9\linewidth}{!}{$
\mathcal{L}_{sn-dist} = \frac{1}{N_{T}\times r+N_{S}} \sum\nolimits_{\ast, \ast'\in\{i\cup jl\}} \mathcal{L}_{ICT} (\ast, \ast').
$}
\end{equation}
It is possible that a snippet pair will include two snippets from the same target video. In such case, the corresponding $\mathcal{L}_{ICT}$ across the snippet pair can be viewed as a low-ordered cross-snippet consistency loss. This implies that optimizing $\mathcal{L}_{cross}$ and $\mathcal{L}_{sn-dist}$ share the common goal of improving feature discriminability for more effective video domain adaptation.

\noindent \textbf{Statistical Snippet Alignment.}
To improve the stability of snippet-level alignment, we adopt a statistical alignment strategy apart from the aforementioned semantic alignment strategies. The statistical alignment is performed by minimizing the snippet-level distribution discrepancies $\mathcal{L}_{sn-stat}$ formulated as metrics such as MMD~\cite{long2015learning}, CORAL~\cite{sun2016return}, and MDD~\cite{zhang2019bridging}. Compared to the adversarial-based adaptation strategy more commonly used in prior VUDA tasks~\cite{chen2019temporal,xu2022aligning,choi2020shuffle}, minimizing discrepancies does not require additional network structures (e.g., domain classifiers), thus is more stable. The MDD~\cite{zhang2019bridging} metric is empirically selected. The overall optimization loss function for FSVDA is therefore:
\begin{equation}
\label{eqn:method:overall-loss}
\resizebox{.9\linewidth}{!}{$
\mathcal{L} = \mathcal{L}_{pred} + \lambda_{sem}(\mathcal{L}_{proto} + \mathcal{L}_{cross} + \mathcal{L}_{sn-dist}) + \lambda_{stat}\mathcal{L}_{sn-stat},
$}
\end{equation}
where $\lambda_{sem}$ and $\lambda_{stat}$ are the tradeoff hyper-parameters for the semantic and statistical snippet alignment losses.

\begin{algorithm}[!t]
\caption{Training with SSA\textsuperscript{2}lign for FSVDA}
\scriptsize
\begin{algorithmic}[]
\REQUIRE $\mathcal{D}_{S}=\{(V_{S, i},y_{S, i})\}^{N_S}_{i=1}$, $\mathcal{D}_{T}=\{(V_{T, j},y_{T, j})\}^{N_T}_{i=1}$, $N_T\ll N_S$.
\WHILE{Training}
\STATE Obtain $r$ target snippets $s_{T, jl}$ from $V_{T, j}$ and one source snippet $s_{S, i}$ from $V_{S, i}$ via SSA.
\STATE Obtain features $\mathbf{f}_{S, i}$, $\mathbf{f}_{T, jl}$, predictions ${o}_{S, i}$, ${o}_{T, jl}$.
\STATE Compute prediction loss as Eq. \ref{eqn:method:ce-loss-all}.
\STATE Obtain snippet attention as Eq. \ref{eqn:method:snippet-attn-w} and normalize. Update $\mathbf{f}_{T, jl}$ to $\mathbf{f'}_{T, jl}$.
\IF{epoch $>e$}
\STATE Obtain target prototypes $Pr_{x}$ as Eq. \ref{eqn:method:proto-tgt}.-\ref{eqn:method:proto-update-tgt}.
\STATE Compute prototype alignment loss as Eq. \ref{eqn:method:proto-align-loss}.
\ENDIF
\STATE Compute cross-snippet consistency loss as Eq. \ref{eqn:method:const-cross-loss}.
\STATE Compute snippet distribution loss as Eq. \ref{eqn:method:ict-tgt-one}-\ref{eqn:method:snippet-dist-tgt}.
\STATE Compute and optimize overall loss as Eq. \ref{eqn:method:overall-loss}.
\ENDWHILE
\ENSURE Trained feature extractor $Time$ and classifier $H$.
\end{algorithmic}
\label{algorithm:method:ssa2lign}
\end{algorithm}

\begin{table*}[!ht]
\center
\resizebox{1.\linewidth}{!}{\noindent
\begin{tabular}{l|l|cccccccccccca|cccccca}
\hline
\hline
  \multirow{2}{*}{Methods} &
  \multirow{2}{*}{Publication} &
  \multicolumn{13}{c|}{\textbf{Daily-DA}} &
  \multicolumn{7}{c}{\textbf{Sports-DA}} \Tstrut\Bstrut\\
\cline{3-22}
& & H$\to$A & M$\to$A & KD$\to$A & A$\to$H & M$\to$H & KD$\to$H & H$\to$M & A$\to$M & KD$\to$M & M$\to$KD & H$\to$KD & A$\to$KD & Avg. & KS$\to$U & S$\to$U & U$\to$S & KS$\to$S & U$\to$KS & S$\to$KS & Avg.\\
\hline
TSF & -
& 37.859 & 32.584 & 31.110 & 44.583 & 57.083 & 45.833 & 36.500 & 30.000 & 34.500 & 61.656 & 58.897 & 75.724 & 45.527 & 91.657 & 91.069 & 76.368 & 77.737 & 87.768 & 85.118 & 84.953\\
TSF w/ T & -
& 39.565 & 39.488 & 39.410 & 61.667 & 62.917 & 62.500 & 41.500 & 38.750 & 36.500 & 77.793 & 80.276 & 83.586 & 55.329 & 92.480 & 93.420 & 78.947 & 79.052 & 88.021 & 87.003 & 86.487\\
\hline
TRX & CVPR-21
& 31.420 & 31.420 & 31.032 & 42.083 & 49.166 & 44.000 & 31.250 & 30.000 & 26.750 & 69.104 & 73.103 & 65.517 & 43.737 & 87.074 & 86.487 & 76.947 & 73.474 & 83.129 & 83.762 & 81.812\\
STRM & CVPR-22
& 33.825 & 32.351 & 32.894 & 43.333 & 50.833 & 44.417 & 30.750 & 29.500 & 28.250 & 72.138 & 74.620 & 68.965 & 45.156 & 91.539 & 90.012 & 78.579 & 75.158 & 86.901 & 84.628 & 84.470\\
HyRSM & CVPR-22
& 38.092 & 35.377 & 33.747 & 45.833 & 54.583 & 48.167 & 33.750 & 31.500 & 29.500 & 75.172 & 76.137 & 70.344 & 47.684 & 92.714 & 90.717 & 79.526 & 76.684 & 87.054 & 84.883 & 85.263\\
\hline
DANN & ICML-15
& 37.471 & 39.721 & 38.557 & 65.417 & 61.667 & 55.833 & 43.750 & 41.250 & 42.000 & 73.655 & 79.173 & 83.173 & 55.139 & 93.067 & 92.127 & 79.211 & 81.316 & 85.525 & 88.634 & 86.647\\
MK-MMD & ICML-15
& 35.299 & 42.746 & 35.609 & 64.167 & 63.333 & 56.667 & 44.000 & 41.750 & 36.500 & 76.690 & 81.931 & 79.862 & 54.879 & 92.597 & 93.420 & 80.737 & 77.842 & 84.760 & 88.124 & 86.247\\
MDD & ICML-19
& 42.514 & 42.281 & 42.901 & 64.583 & 64.167 & 57.917 & 45.000 & 39.500 & 37.750 & 75.173 & 81.517 & 84.276 & 56.465 & 93.184 & 93.067 & 78.474 & 79.105 & 86.697 & 87.716 & 86.374\\
SAVA & ECCV-20
& 39.178 & 41.660 & 41.738 & 63.333 & 63.333 & 60.000 & 42.750 & 41.500 & 39.250 & 77.517 & 81.242 & 80.690 & 56.016 & 93.302 & 91.540 & 79.263 & 80.474 & 87.614 & 87.512 & 86.617\\
ACAN & TNNLS-22
& 43.832 & 43.755 & 43.677 & 65.417 & 66.667 & 66.250 & 45.750 & 43.000 & 40.750 & 82.483 & 84.966 & 84.414 & 59.247 & 95.770 & 96.710 & 80.158 & 80.263 & 88.327 & 88.583 & 88.302\\
\hline
FADA & NeurIPS-17
& 39.100 & 42.126 & 32.351 & 46.250 & 58.750 & 47.500 & 37.250 & 30.750 & 35.250 & 77.241 & 81.103 & 77.517 & 50.432 & 93.655 & 93.655 & 76.947 & 78.316 & 88.736 & 86.086 & 86.233\\
d-SNE & CVPR-19
& 41.583 & 44.065 & 38.014 & 67.083 & 65.417 & 61.667 & 44.500 & 43.250 & 41.000 & 78.759 & 82.759 & 83.448 & 57.629 & 95.417 & 94.830 & 81.105 & 82.316 & 89.755 & 83.509 & 87.822\\
\hline
\textbf{SSA\textsuperscript{2}lign} & -
& \textbf{52.133} & \textbf{52.211} & \textbf{51.746} & \textbf{78.333} & \textbf{75.417} & \textbf{74.583} & \textbf{47.750} & \textbf{46.750} & \textbf{48.250} & \textbf{84.690} & \textbf{86.483} & \textbf{89.655} & \textbf{65.667} & \textbf{98.589} & \textbf{98.237} & \textbf{87.263} & \textbf{88.105} & \textbf{92.966} & \textbf{93.017} & \textbf{93.029}\Tstrut\Bstrut\\
\hline
\hline
\end{tabular}
}
\vspace{-2pt}
\caption{Results for 10-shot ($k=10$) FSVDA on Daily-DA and Sports-DA.}
\label{table:4-1-sota-k10-1}
\vspace{-6pt}
\end{table*}

\begin{table*}[!ht]
\center
\resizebox{1.\linewidth}{!}{\noindent
\begin{tabular}{l|l|cccccccccccca|cccccca}
\hline
\hline
  \multirow{2}{*}{Methods} &
  \multirow{2}{*}{Publication} &
  \multicolumn{13}{c|}{\textbf{Daily-DA}} &
  \multicolumn{7}{c}{\textbf{Sports-DA}} \Tstrut\Bstrut\\
\cline{3-22}
& & H$\to$A & M$\to$A & KD$\to$A & A$\to$H & M$\to$H & KD$\to$H & H$\to$M & A$\to$M & KD$\to$M & M$\to$KD & H$\to$KD & A$\to$KD & Avg. & KS$\to$U & S$\to$U & U$\to$S & KS$\to$S & U$\to$KS & S$\to$KS & Avg.\\
\hline
TSF & -
& 37.859 & 32.584 & 31.110 & 44.583 & 57.083 & 45.833 & 36.500 & 30.000 & 34.500 & 61.656 & 58.897 & 75.724 & 45.527 & 91.657 & 91.069 & 76.368 & 77.737 & 87.768 & 85.118 & 84.953\\
TSF w/ T & -
& 40.186 & 40.031 & 37.083 & 60.043 & 60.043 & 52.960 & 34.750 & 36.000 & 33.250 & 79.448 & 66.207 & 69.103 & 50.759 & 91.892 & 93.302 & 78.736 & 78.315 & 87.971 & 86.799 & 86.169\\
\hline
TRX & CVPR-21
& 32.794 & 30.260 & 28.987 & 39.425 & 47.446 & 40.349 & 29.000 & 27.750 & 24.750 & 69.545 & 63.032 & 55.882 & 40.768 & 86.531 & 86.955 & 76.342 & 70.946 & 81.827 & 80.411 & 80.502\\
STRM & CVPR-22
& 35.318 & 32.512 & 30.979 & 40.150 & 47.494 & 39.300 & 27.250 & 26.250 & 26.500 & 72.489 & 62.580 & 57.777 & 41.550 & 91.003 & 90.017 & 77.844 & 73.494 & 86.378 & 82.140 & 83.479\\
HyRSM & CVPR-22
& 39.065 & 35.088 & 31.061 & 42.736 & 52.098 & 43.740 & 31.000 & 29.250 & 28.000 & 75.646 & 63.889 & 58.598 & 44.181 & 92.166 & 90.769 & 79.122 & 75.272 & 86.251 & 81.713 & 84.215\\
\hline
DANN & ICML-15
& 40.496 & 38.789 & 36.385 & 60.833 & 58.750 & 52.917 & 41.750 & 38.500 & 39.750 & 74.345 & 66.759 & 69.655 & 51.577 & 92.245 & 93.067 & 78.421 & 75.631 & 85.117 & 82.161 & 84.440\\
MK-MMD & ICML-15
& 38.867 & 43.910 & 34.600 & 58.333 & 57.083 & 54.583 & 42.250 & 35.000 & 35.500 & 75.311 & 68.138 & 69.380 & 51.080 & 92.010 & 93.184 & 78.737 & 75.000 & 84.505 & 83.486 & 84.487\\
MDD & ICML-19
& 41.893 & 42.669 & 38.402 & 61.250 & 62.500 & 55.417 & 43.250 & 40.000 & 38.500 & 75.724 & 68.552 & 70.896 & 53.254 & 92.715 & 92.597 & 79.105 & 79.790 & 86.544 & 83.588 & 85.723\\
SAVA & ECCV-20
& 40.962 & 37.238 & 38.247 & 60.000 & 62.917 & 55.833 & 40.750 & 38.750 & 35.250 & 77.793 & 67.448 & 68.965 & 52.013 & 92.480 & 92.832 & 78.053 & 76.211 & 83.129 & 81.702 & 84.068\\
ACAN & TNNLS-22
& 44.453 & 44.298 & 41.350 & 63.333 & 63.333 & 56.250 & 39.000 & 40.250 & 37.500 & 80.552 & 70.897 & 73.793 & 54.584 & 95.182 & 96.592 & 79.947 & 79.526 & 88.175 & 88.379 & 87.967\\
\hline
FADA & NeurIPS-17
& 40.747 & 41.584 & 30.065 & 43.034 & 55.534 & 41.058 & 33.000 & 27.000 & 32.500 & 77.792 & 67.185 & 64.736 & 46.186 & 93.009 & 93.860 & 76.040 & 76.173 & 87.502 & 83.375 & 84.993\\
d-SNE & CVPR-19
& 41.994 & 44.162 & 35.738 & 64.927 & 63.365 & 56.936 & 41.250 & 41.750 & 39.750 & 79.448 & 72.530 & 73.296 & 54.596 & 94.992 & 94.734 & 80.983 & 81.768 & 89.371 & 81.749 & 87.266\\
\hline
\textbf{SSA\textsuperscript{2}lign} & -
& \textbf{52.366} & \textbf{51.978} & \textbf{47.401} & \textbf{76.667} & \textbf{72.917} & \textbf{70.417} & \textbf{47.000} & \textbf{46.250} & \textbf{47.500} & \textbf{86.759} & \textbf{79.310} & \textbf{81.793} & \textbf{63.363} & \textbf{97.062} & \textbf{97.885} & \textbf{84.053} & \textbf{86.211} & \textbf{91.182} & \textbf{90.214} & \textbf{91.101}\Tstrut\Bstrut\\
\hline
\hline
\end{tabular}
}
\vspace{-2pt}
\caption{Results for 5-shot ($k=5$) FSVDA on Daily-DA and Sports-DA.}
\label{table:4-2-sota-k5-2}
\vspace{-6pt}
\end{table*}

\begin{table*}[!ht]
\center
\resizebox{1.\linewidth}{!}{\noindent
\begin{tabular}{l|l|cccccccccccca|cccccca}
\hline
\hline
  \multirow{2}{*}{Methods} &
  \multirow{2}{*}{Publication} &
  \multicolumn{13}{c|}{\textbf{Daily-DA}} &
  \multicolumn{7}{c}{\textbf{Sports-DA}} \Tstrut\Bstrut\\
\cline{3-22}
& & H$\to$A & M$\to$A & KD$\to$A & A$\to$H & M$\to$H & KD$\to$H & H$\to$M & A$\to$M & KD$\to$M & M$\to$KD & H$\to$KD & A$\to$KD & Avg. & KS$\to$U & S$\to$U & U$\to$S & KS$\to$S & U$\to$KS & S$\to$KS & Avg.\\
\hline
TSF & -
& 37.859 & 32.584 & 31.110 & 44.583 & 57.083 & 45.833 & 36.500 & 30.000 & 34.500 & 61.656 & 58.897 & 75.724 & 45.527 & 91.657 & 91.069 & 76.368 & 77.737 & 87.768 & 85.118 & 84.953\\
TSF w/ T & -
& 37.937 & 34.135 & 33.049 & 51.250 & 58.750 & 46.667 & 37.750 & 34.500 & 35.250 & 74.069 & 60.414 & 63.724 & 47.291 & 91.165 & 92.832 & 75.368 & 76.578 & 86.595 & 85.372 & 84.652\\
\hline
TRX & CVPR-21
& 24.679 & 23.331 & 25.059 & 32.052 & 43.341 & 30.229 & 28.000 & 27.500 & 23.250 & 66.187 & 52.086 & 49.378 & 35.424 & 84.898 & 86.300 & 71.852 & 69.974 & 80.815 & 79.784 & 78.937\\
STRM & CVPR-22
& 28.037 & 24.181 & 26.924 & 33.853 & 45.022 & 30.038 & 25.750 & 26.000 & 25.000 & 69.137 & 52.130 & 50.492 & 36.380 & 89.640 & 89.176 & 73.719 & 72.496 & 85.682 & 81.630 & 82.057\\
HyRSM & CVPR-22
& 30.939 & 27.142 & 26.970 & 35.670 & 48.420 & 34.360 & 30.000 & 28.750 & 26.500 & 71.710 & 52.927 & 50.582 & 38.664 & 90.679 & 90.241 & 74.732 & 73.273 & 84.706 & 81.047 & 82.446\\
\hline
DANN & ICML-15
& 30.566 & 28.627 & 34.057 & 53.750 & 51.667 & 42.083 & 39.750 & 37.250 & 33.000 & 73.655 & 52.966 & 64.276 & 45.137 & 91.637 & 91.422 & 72.895 & 76.895 & 84.709 & 82.545 & 83.350\\
MK-MMD & ICML-15
& 29.403 & 32.506 & 31.963 & 54.583 & 55.417 & 44.167 & 38.500 & 37.000 & 33.750 & 72.000 & 56.000 & 63.035 & 45.694 & 90.717 & 91.070 & 74.684 & 74.211 & 85.830 & 84.047 & 83.426\\
MDD & ICML-19
& 31.652 & 33.592 & 34.523 & 54.167 & 56.667 & 47.083 & 42.250 & 38.500 & 34.750 & 70.621 & 56.138 & 59.448 & 46.616 & 91.422 & 92.715 & 74.369 & 74.895 & 82.823 & 82.925 & 83.191\\
SAVA & ECCV-20
& 31.031 & 33.436 & 32.971 & 50.833 & 58.333 & 42.917 & 40.500 & 39.750 & 37.750 & 72.138 & 55.724 & 62.069 & 46.454 & 89.307 & 92.832 & 73.211 & 73.842 & 81.753 & 80.377 & 81.887\\
ACAN & TNNLS-22
& 38.635 & 35.609 & 35.299 & 55.000 & 61.667 & 46.667 & 38.250 & 38.750 & 35.750 & 76.276 & 59.311 & 62.621 & 48.653 & 93.750 & 96.240 & 75.368 & 77.052 & 85.658 & 87.105 & 85.862\\
\hline
FADA & NeurIPS-17
& 33.881 & 34.136 & 25.565 & 35.523 & 53.232 & 31.256 & 32.500 & 27.000 & 32.750 & 73.861 & 57.150 & 57.908 & 41.230 & 91.353 & 93.126 & 71.829 & 75.042 & 86.750 & 82.508 & 83.435\\
d-SNE & CVPR-19
& 36.263 & 37.859 & 32.131 & 59.195 & 60.914 & 50.006 & 40.500 & 41.000 & 38.500 & 76.293 & 64.298 & 62.934 & 49.991 & 93.896 & 94.492 & 76.440 & 77.111 & 87.857 & 81.049 & 85.141\\
\hline
\textbf{SSA\textsuperscript{2}lign} & -
& \textbf{44.831} & \textbf{46.780} & \textbf{45.306} & \textbf{68.750} & \textbf{70.833} & \textbf{62.083} & \textbf{46.750} & \textbf{46.500} & \textbf{45.000} & \textbf{79.724} & \textbf{65.793} & \textbf{71.586} & \textbf{57.828} & \textbf{96.592} & \textbf{97.415} & \textbf{80.053} & \textbf{80.947} & \textbf{88.940} & \textbf{89.755} & \textbf{88.950}\Tstrut\Bstrut\\
\hline
\hline
\end{tabular}
}
\vspace{-2pt}
\caption{Results for 3-shot ($k=3$) FSVDA on Daily-DA and Sports-DA.}
\label{table:4-3-sota-k3-3}
\vspace{-6pt}
\end{table*}

\noindent \textbf{Snippet Attention.}
With multiple snippets leveraged per target video for both semantic and statistical snippet alignments, it is unreasonable to leverage each snippet equally since it is intuitive that the importance of each target snippet is uneven. We thus propose a snippet attention to weigh the impact of different target snippets on the domain alignment dynamically. Intuitively, a snippet whose output prediction is the most accurate, i.e., whose classification is closest to its given ground truth, should be focused during alignment. A simple yet effective expression of how accurate the snippet's output prediction is would be the inverse of the cross-entropy loss. The snippet attention weights are therefore built upon the inverse of the cross-entropy loss of the snippet, along with a residual connection for more stable optimization, expressed as:
\begin{equation}
\label{eqn:method:snippet-attn-w}
w_{jl} = 1 + \frac{1}{l_{ce}({o}_{T, jl}, y_{j, T})}.
\end{equation}
The snippet attention weights are subsequently normalized across the $r$ snippets corresponding to the same target video, expressed as $\overline{w}_{jl} = w_{jl}/\frac{1}{r}\sum\nolimits_{l'=1}^{r} w_{jl'}.$ The normalized snippet attention weight $\overline{w}_{jl}$ is then applied to the target snippet features, forming the weighted target snippet features by $\mathbf{f'}_{T, jl} = \overline{w}_{jl}\mathbf{f}_{T, jl}$, which are then aligned with the source domain through the semantic and statistical snippet alignments by replacing the features $\mathbf{f}_{T, jl}$ with $\mathbf{f'}_{T, jl}$.

\noindent \textbf{SSA\textsuperscript{2}lign.}
Finally, we sum up our proposed SSA\textsuperscript{2}lign in Algorithm \ref{algorithm:method:ssa2lign}. The snippet features, SSA, and snippet attention are leveraged only during training. During testing, target video representations are obtained by uniform sampling across the target testing videos, while the video features and their output predictions are obtained by directly applying the trained feature extractor and classifier to the uniformly sampled target video representations.

\section{Experiments}
\label{section:exps}

In this section, we evaluate our proposed SSA\textsuperscript{2}lign across two challenging cross-domain action recognition benchmarks: Daily-DA and Sports-DA~\cite{xu2023multi}, which cover a wide range of cross-domain scenarios. We present superior results on both benchmarks. Further, ablation studies and analysis of SSA\textsuperscript{2}lign are also presented to justify the design of SSA\textsuperscript{2}lign.

\subsection{Experimental Settings}
\label{section:exps:setting}
\textbf{Daily-DA} is a challenging dataset that has been leveraged in prior VUDA works~\cite{xu2023multi,xu2022source,xu2022extern}. It covers both normal and low-illumination videos and is constructed from four datasets: ARID (A)~\cite{xu2021arid}, HMDB51 (H)~\cite{kuehne2011hmdb}, Moments-in-Time (M)~\cite{monfort2019moments}, and Kinetics-600 (KD)~\cite{carreira2018short}. HMDB51, Moments-in-Time, and Kinetics-600 are widely used for action recognition benchmarking, while ARID is a recent dark dataset, with videos shot under adverse illumination. Daily-DA contains 18,949 videos from 8 classes, with 12 cross-domain action recognition tasks. \textbf{Sports-DA} is a large-scale cross-domain video dataset, built from UCF101 (U), Sports-1M (S)~\cite{karpathy2014large}, and Kinetics-600 (KS), with 40,718 videos from 23 action classes, and includes 6 cross-domain action recognition tasks. Refer to prior FSDA/FSVDA works~\cite{gao2020pairwise-tip,gao2020pairwise-tnnls,gao2022acrofod}, we evaluate SSA\textsuperscript{2}lign on both benchmarks with $k=(3,5,10)$ target videos per action class (i.e., 3-shot, 5-shot and 10-shot VDA tasks).

For a fair comparison, all methods examined and experiments conducted in this section adopt the TimeSFormer~\cite{zhou2018temporal} as the feature extractor, pre-trained on Kinetics-400~\cite{kay2017kinetics}. All experiments are implemented with the PyTorch~\cite{paszke2019pytorch} library. We set the length of snippets and the number of snippets per target video via SSA empirically as $m=8, r=3$. Hyper-parameters $\lambda_{sem}$, $\lambda_{stat}$ and $\lambda_{P}$ are empirically set to $1.0$, $1.0$, and $0.6$ and are fixed. 
\textit{More specifications on benchmark details and network implementation are provided in the appendix}.

\begin{table*}[!ht]
\center
\resizebox{.95\linewidth}{!}{\noindent
\begin{tabular}{l|cccccc|ccc|ccc|ccc|cc|cc|cc|c}
\hline
\hline
  \multirow{3}{*}{Methods} &
  \multicolumn{6}{|c|}{\multirow{2}{*}{Components}} &
  \multicolumn{9}{c|}{Daily-DA} & 
  \multicolumn{6}{c|}{Sports-DA} & \cellcolor{LightGray}
  \Tstrut\Bstrut\\
\cline{8-22}
  & & & & & & & 
  \multicolumn{3}{c|}{$k=10$} &
  \multicolumn{3}{c|}{$k=5$} &
  \multicolumn{3}{c|}{$k=3$} &
  \multicolumn{2}{c|}{$k=10$} &
  \multicolumn{2}{c|}{$k=5$} &
  \multicolumn{2}{c|}{$k=3$} & \cellcolor{LightGray}
  \Tstrut\Bstrut\\
\cline{2-22}
& SSA & Sn-Attn & $\mathcal{L}_{proto}$ & $\mathcal{L}_{cross}$ & $\mathcal{L}_{sn-dist}$ & $\mathcal{L}_{sn-stat}$ & H$\to$A & M$\to$A & KD$\to$A & H$\to$A & M$\to$A & KD$\to$A & H$\to$A & M$\to$A & KD$\to$A & U$\to$S & KS$\to$S & U$\to$S & KS$\to$S & U$\to$S & KS$\to$S & 
  \multirow{-3}{*}{\cellcolor{LightGray}Avg.}\Tstrut\\
\hline
\multirow{2}{*}{TSF w/ T}
&  &  &  &  &  &  & 39.565 & 39.488 & 39.410 & 40.186 & 40.031 & 37.083 & 37.937 & 34.135 & 33.049 & 78.947 & 79.052 & 78.736 & 78.315 & 75.368 & 76.578 & \cellcolor{LightGray}53.859\Tstrut\\
& \cmark &  &  &  &  &  & 41.660 & 41.971 & 42.048 & 42.824 & 42.281 & 38.247 & 38.790 & 36.385 & 35.221 & 81.053 & 81.210 & 80.789 & 80.421 & 75.894 & 77.210 & \cellcolor{LightGray}55.734\Tstrut\\
\hline
\multirow{7}{*}{\textbf{SSA\textsuperscript{2}lign}}
&  & \cmark & \cmark & \cmark & \cmark & \cmark & 45.616 & 46.315 & 45.695 & 46.470 & 45.686 & 41.738 & 39.168 & 40.962 & 39.488 & 84.316 & 85.473 & 81.053 & 83.579 & 77.264 & 77.368 & \cellcolor{LightGray}58.679\Tstrut\\
& \cmark &  & \cmark & \cmark & \cmark &  & 51.047 & 51.125 & 50.427 & 50.970 & 50.729 & 46.392 & 43.590 & 45.850 & 44.220 & 86.579 & 87.368 & 83.211 & 85.632 & 79.316 & 80.368 & \cellcolor{LightGray}62.455\Tstrut\\
& \cmark & \cmark & \cmark &  &  &  & 49.883 & 49.806 & 49.729 & 50.349 & 49.255 & 45.074 & 42.193 & 44.530 & 42.901 & 85.579 & 86.737 & 82.737 & 84.737 & 78.685 & 79.473 & \cellcolor{LightGray}61.445\Tstrut\\
& \cmark & \cmark &  & \cmark &  &  & 50.194 & 50.504 & 49.651 & 50.271 & 49.875 & 45.772 & 42.659 & 45.151 & 43.056 & 85.842 & 87.052 & 82.579 & 85.053 & 78.948 & 79.579 & \cellcolor{LightGray}61.746\Tstrut\\
& \cmark & \cmark &  &  & \cmark &  & 48.176 & 48.565 & 46.858 & 47.556 & 48.091 & 43.289 & 40.642 & 42.513 & 41.970 & 84.684 & 85.947 & 81.948 & 83.737 & 77.737 & 78.631 & \cellcolor{LightGray}60.023\Tstrut\\
& \cmark & \cmark & \cmark & \cmark & \cmark &  & 51.357 & 51.435 & 50.893 & 51.823 & 51.427 & 46.548 & 43.900 & 46.004 & 44.918 & 86.684 & 87.631 & 83.632 & 85.685 & 79.685 & 80.421 & \cellcolor{LightGray}62.798\Tstrut\\
& \cmark & \cmark & \cmark & \cmark & \cmark & \cmark & \textbf{52.133} & \textbf{52.211} & \textbf{51.746} & \textbf{52.366} & \textbf{51.978} & \textbf{47.401} & \textbf{44.831} & \textbf{46.780} & \textbf{45.306} & \textbf{87.263} & \textbf{88.105} & \textbf{84.053} & \textbf{86.211} & \textbf{80.053} & \textbf{80.947} & \cellcolor{LightGray}\textbf{63.425}\Tstrut\Bstrut\\
\hline
\hline
\end{tabular}
}
\vspace{2pt}
\caption{Ablation studies of the components of SSA\textsuperscript{2}lign on 5 cross-domain tasks over Daily-DA and Sports-DA.}
\label{table:4-4-ablation}
\vspace{-6pt}
\end{table*}

\begin{table*}[!ht]
\center
\resizebox{.9\linewidth}{!}{\noindent
\begin{tabular}{l|ccc|ccc|ccc|cc|cc|cc|c|c|c|c}
\hline
\hline
  \multirow{3}{*}{Methods} &
  \multicolumn{9}{c|}{Daily-DA} &
  \multicolumn{6}{c|}{Sports-DA} & \cellcolor{LightGray}  & \cellcolor{LightGray} & \cellcolor{LightGray}  & \cellcolor{LightGray}
  \Tstrut\Bstrut\\
\cline{2-16}
  & 
  \multicolumn{3}{c|}{$k=10$} &
  \multicolumn{3}{c|}{$k=5$} &
  \multicolumn{3}{c|}{$k=3$} &
  \multicolumn{2}{c|}{$k=10$} &
  \multicolumn{2}{c|}{$k=5$} &
  \multicolumn{2}{c|}{$k=3$} & \cellcolor{LightGray}  & \cellcolor{LightGray} & \cellcolor{LightGray}  & \cellcolor{LightGray}
  \Tstrut\Bstrut\\
\cline{2-16}
& H$\to$A & M$\to$A & KD$\to$A & H$\to$A & M$\to$A & KD$\to$A & H$\to$A & M$\to$A & KD$\to$A & U$\to$S & KS$\to$S & U$\to$S & KS$\to$S & U$\to$S & KS$\to$S & 
  \multirow{-3}{*}{\cellcolor{LightGray}Avg.} &
  \multirow{-3}{*}{\cellcolor{LightGray}$\Delta$ Avg.} &
  \multirow{-3}{*}{\cellcolor{LightGray}GFLOPS} &
  \multirow{-3}{*}{\cellcolor{LightGray}$\Delta$ GFLOPS} \Tstrut\\
\hline
SSA\textsuperscript{2}lign-CORAL
& 51.900 & 51.978 & 51.513 & 51.978 & 51.582 & 47.091 & 44.521 & 46.392 & 45.073 & 87.052 & 87.842 & 83.895 & 86.000 & 79.895 & 80.684 & \cellcolor{LightGray}63.160 & \cellcolor{LightGray}-0.265 & \cellcolor{LightGray}1302 & \cellcolor{LightGray}-8 \Tstrut\\
SSA\textsuperscript{2}lign-MMD
& 51.668 & 51.901 & 51.281 & 51.978 & 51.505 & 47.013 & 44.521 & 46.315 & 44.841 & 87.052 & 87.842 & 83.842 & 85.895 & 79.790 & 80.631 & \cellcolor{LightGray}63.072 & \cellcolor{LightGray}-0.353 & \cellcolor{LightGray}1312 & \cellcolor{LightGray}+2 \Tstrut\\
SSA\textsuperscript{2}lign-FC
& 52.831 & 52.909 & 52.367 & 52.987 & 52.358 & 47.556 & 45.296 & 47.245 & 45.306 & 87.158 & 88.263 & 84.527 & 86.685 & 79.790 & 81.263 & \cellcolor{LightGray}63.769 & \cellcolor{LightGray}+0.344 & \cellcolor{LightGray}1472 & \cellcolor{LightGray}+162 \Tstrut\\
SSA\textsuperscript{2}lign-SP
& 50.969 & 51.202 & 50.970 & 51.280 & 51.272 & 46.470 & 43.822 & 45.616 & 44.685 & 86.895 & 87.421 & 83.369 & 85.737 & 79.264 & 80.473 & \cellcolor{LightGray}62.630 & \cellcolor{LightGray}-0.795 & \cellcolor{LightGray}1390 & \cellcolor{LightGray}+80 \Tstrut\\
SSAlign (w/ spatial aug.)
& 45.383 & 46.703 & 45.230 & 45.772 & 45.841 & 41.117 & 39.633 & 40.264 & 38.635 & 83.527 & 85.631 & 80.579 & 83.368 & 76.948 & 77.736 & \cellcolor{LightGray}58.424 & \cellcolor{LightGray}-5.001 & \cellcolor{LightGray}1325 & \cellcolor{LightGray}+15 \Tstrut\\
\hline
SSA\textsuperscript{2}lign
& 52.133 & 52.211 & 51.746 & 52.366 & 51.978 & 47.401 & 44.831 & 46.780 & 45.306 & 87.263 & 88.105 & 84.053 & 86.211 & 80.053 & 80.947 & \cellcolor{LightGray}63.425 & \cellcolor{LightGray}- & \cellcolor{LightGray}1310 & \cellcolor{LightGray}- \Tstrut\Bstrut\\
\hline
\hline
\end{tabular}
}
\vspace{2pt}
\caption{Ablation studies of the alignment details of SSA\textsuperscript{2}lign on 5 cross-domain tasks over Daily-DA and Sports-DA.}
\label{table:4-5-ablation-2}
\vspace{-12pt}
\end{table*}

\subsection{Overall Results and Comparisons}
\label{section:exps:results}
We compare SSA\textsuperscript{2}lign with state-of-the-art FSDA approaches, and prevailing UDA/VUDA and few-shot action recognition (FSAR) approaches. These methods include: FADA~\cite{motiian2017few}, d-SNE~\cite{xu2019dsne} designed for image-based FSDA; DANN~\cite{ganin2015unsupervised}, MK-MMD~\cite{long2015learning}, MDD~\cite{zhang2019bridging}, SAVA~\cite{choi2020shuffle} and ACAN~\cite{xu2022aligning}, designed for UDA/VUDA; and TRX~\cite{perrett2021temporal}, STRM~\cite{thatipelli2022spatio}, and HyRSM~\cite{wang2022hybrid} proposed for FSAR. To adapt the FSAR approaches for FSVDA, the source domain is used for meta-training and the target domain is used for the meta-testing, while target labels are available for optimizing the cross-entropy loss to adapt UDA/VUDA approaches for FSVDA. We also report the results of the source-only model (denoted as TSF) by applying the model trained with only source data directly to the target data; and the source with few-shot target model (denoted as TSF w/ T) by optimizing only the prediction loss $\mathcal{L}_{pred}$ for training. We report the top-1 accuracy on the target domains, averaged on 5 different settings of available target data randomly selected and each with 5 runs (25 runs in total). Tables \ref{table:4-1-sota-k10-1}-\ref{table:4-3-sota-k3-3} show comparison of SSA\textsuperscript{2}lign against the above methods.

Results in Tables \ref{table:4-1-sota-k10-1}-\ref{table:4-3-sota-k3-3} show that the novel SSA\textsuperscript{2}lign achieves the state-of-the-art results on all 18 cross-domain action recognition tasks across both cross-domain benchmarks, outperforming prior UDA/VUDA, FSDA or FSAR approaches by noticeable margins. Notably, SSA\textsuperscript{2}lign outperforms all prior FSDA approaches originally designed for image-based FSDA (i.e., FADA and d-SNE) consistently on all tasks, by a relative average of $13\%$ over the second-best performances on Daily-DA (across 3 $k$-shot settings and 12 tasks), and a relative average of $4.2\%$ on Sports-DA (across 3 $k$-shot settings and 6 tasks). The consistent improvements justify empirically the effectiveness of augmenting and aligning both semantic information and statistical distribution at the snippet level for FSVDA.

It is also observed that prior FSDA and UDA/VUDA methods could not perform well on FSVDA tasks. Notably, even when $k=10$ target videos are available per class, all but one of the evaluated FSDA and UDA/VUDA approaches result in performances inferior to that trained with only $\mathcal{L}_{pred}$ without any adaptation (i.e., TSF w/ T). Prior FSDA approaches do not incorporate temporal features and their related semantic information, which are crucial for tackling FSVDA, while UDA/VUDA methods are not effective when target information is not fully available. Negative improvements are more severe when $k$ decreases. It is also noted that at small $k$ values (e.g., $k=3$), the performance of TSF w/ T could be inferior to that trained without target data (i.e., TSF). This suggests that the few target data could be outliers of the target domain, whose distribution differs greatly from the other target data, resulting in a severe negative impact. Prior FSAR approaches could not tackle FSVDA as well, producing even poorer results than all UDA/VUDA approaches examined. This can be caused by domain shift that exists between data for the meta-training and meta-testing. Feature extractors trained via meta-training on the source domain could not be simply applied to the meta-testing phase on the target domain.

\begin{figure}[!t]
\begin{center}
  \includegraphics[width=.8\linewidth]{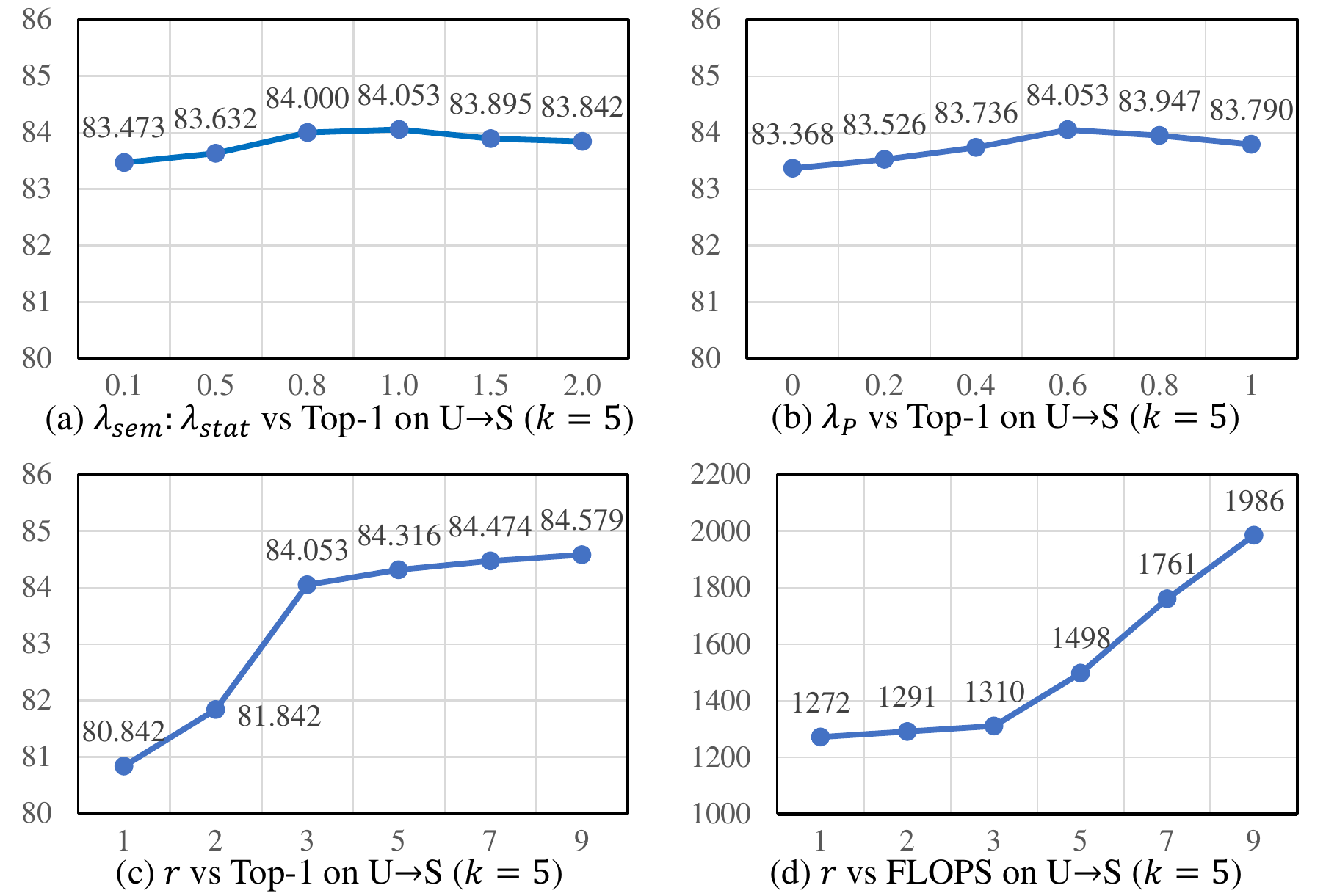}
\end{center}
\vspace{-12pt}
\caption{Sensitivity of hyper-parameters on U$\to$S task.}
\label{figure:4-2-hyperparameter}
\end{figure}

\begin{figure*}[!ht]
\begin{center}
   \includegraphics[width=.75\linewidth]{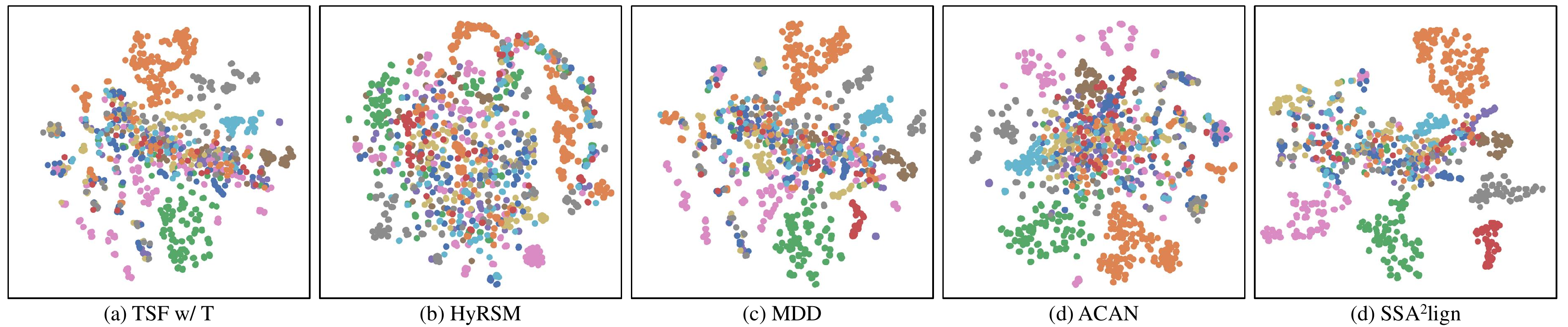}
\end{center}
\vspace{-12pt}
\caption{t-SNE visualizations of target features from (a) TSF w/T, (b) HyRSM, (c) MDD, (d) SSA\textsuperscript{2}lign. Colors denote classes.}
\label{figure:4-3-tsne}
\vspace{-15pt}
\end{figure*}

\subsection{Ablation Studies and Analysis}
\label{section:exps:ablation-analysis}
To gain a comprehensive understanding of SSA\textsuperscript{2}lign and justify its design, we perform extensive ablation studies as in Tables \ref{table:4-4-ablation}-\ref{table:4-5-ablation-2}. The ablation studies explore the effects brought by its components, namely the semantic and statistical alignments, the SSA, and the snippet attention. It further validates the alignment details by assessing against 5 variants: SSA\textsuperscript{2}lign-CORAL and SSA\textsuperscript{2}lign-MMD formulate $\mathcal{L}_{sn-stat}$ as CORAL~\cite{sun2016return} and MDD~\cite{zhang2019bridging}; SSA\textsuperscript{2}lign-FC computes $\mathcal{L}_{cross}$ over all $r\times(r-1)$ snippet pairs for the same target video; SSA\textsuperscript{2}lign-SP minimizes the distance between target snippet features and source class prototypes for $\mathcal{L}_{proto}$; SSAlign (w/ spatial aug.) augments target domain through random spatial augmentation across the frames of $r$ snippets. The ablation studies are conducted on 5 tasks over Daily-DA and Sports-DA. If SSA is not applied, we sample $r$ snippets sequentially from the $1^{st}$ frame of each target video and remain unchanged during training. 

\noindent \textbf{Semantic Alignment.}
As shown in Table \ref{table:4-4-ablation}, with only snippet-level semantic alignment (whether in full or any one of the three perspectives), the performance still surpasses all previous FSDA and UDA/VUDA methods compared. This conforms to our motivation that applying semantic alignment could tackle FSVDA more effectively. Moreover, statistical alignment and snippet attention further improve SSA\textsuperscript{2}lign, but only by a marginal degree.

\noindent \textbf{Superiority of SSA.}
Notably, a significant performance drop is observed when SSA is not applied, which proves the importance of expanding target domain data through SSA for subsequent alignment. The importance of SSA is further verified when we apply SSA for training with augmented snippets but without adaptation which shows a noticeable gain compared to the original TSF w/ T. Further, the significantly inferior performance of SSAlign (w/ spatial aug.) as shown in Table \ref{table:4-5-ablation-2} conforms with the motivation of SSA, which aims for more effective target video domain augmentation while spatial augmentation may undermine temporal correlation across sequential frames. 

\noindent \textbf{Alignment Methods.} 
Table \ref{table:4-5-ablation-2} shows that while formulating $\mathcal{L}_{sn-stat}$ as MDD~\cite{zhang2019bridging} brings the best performance, selecting other metrics brings negligible impact. Further, computing $\mathcal{L}_{cross}$ with all target snippet pairs only brings trivial performance gain at a cost of significant computation overhead ($12\%$ computation increase for $0.54\%$ gain). Further, matching target snippet features to source class prototypes for $\mathcal{L}_{proto}$ results in a performance drop with more computation. The inferior performance could be due to outliers in the source domain which could affect source class prototypes, bringing in source noise that should not be aligned.

\noindent \textbf{Hyper-parameter Sensitivity.}
We focus on studying the sensitivity of $\lambda_{sem}$ and $\lambda_{stat}$ which control the strength of the semantic and statistical snippet alignment losses, $\lambda_{P}$ which relates to the update of target prototypes and $r$ the number of snippets per target video. Without loss of generality, we fix $\lambda_{stat}=1.0$ and study the ratio $\lambda_{sem}\!:\!\lambda_{stat}$ in the range of 0.1 to 1.5. $\lambda_{P}$ is in the range of 0 to 1 which corresponds to using only the initial prototypes or the updated prototypes, and $r$ is in the range of 1 to 9. As shown in Fig. \ref{figure:4-2-hyperparameter}, SSA\textsuperscript{2}lign is robust to ratio $\lambda_{sem}\!:\!\lambda_{stat}$ and $\lambda_{P}$, falling within a margin of $0.683\%$, with the best results obtained at the current default where $\lambda_{sem}\!:\!\lambda_{stat}=1.0$ and $\lambda_{P}=0.6$. SSA\textsuperscript{2}lign is also robust to $r$ when $r\geqslant3$, i.e., when there are multiple snippets obtained via SSA per target video. $r=3$ is selected as significant computation overhead would occur for $r>3$ with marginal gain. Notably, SSA\textsuperscript{2}lign cannot perform when $r<3$, especially when $r=1$ where the $\mathcal{L}_{cross}$ does not work and the target domain is not expanded.

\noindent \textbf{Feature Visualization.}
We further understand the characteristics of SSA\textsuperscript{2}lign by plotting the t-SNE embeddings~\cite{van2008visualizing} of target features with class information from the model trained without adaptation (TSF w/T), HyRSM, MDD, ACAN and SSA\textsuperscript{2}lign for U$\to$S with $k=10$ in Fig. \ref{figure:4-3-tsne}. It is observed that target features from SSA\textsuperscript{2}lign are more clustered and discriminable, corresponding to better performance. Such observation intuitively proves that video domain adaptation can be improved when feature extractors possess stronger discriminability.
However, SSA\textsuperscript{2}lign is not designed to deal explicitly with classes that could be similar spatially or temporally, thus certain features observe lower discriminability, which denotes future work.

\section{Conclusion}
\label{section:concl}

In this work, we propose a novel SSA\textsuperscript{2}lign to tackle the challenging yet realistic Few-Shot Video Domain Adaptation (FSVDA), where only limited labeled target data are available. Without sufficient target data, SSA\textsuperscript{2}lign tackles FSVDA at the snippet level via a simple SSA augmentation and performing the semantic and statistical alignments attentively, where the semantic alignment is further achieved from three perspectives based on semantic information within and across snippets. Extensive experiments and detailed ablation studies across cross-domain action recognition benchmarks validate the superiority of SSA\textsuperscript{2}lign in addressing FSVDA.

\section*{Appendix}
\label{section:appendix}

\textit{This appendix presents more details of the proposed Snippet-attentive Semantic-statistical Alignment with Stochastic Sampling Augmentation (SSA\textsuperscript{2}lign) and is organized as follows: first, we introduce the detailed implementation of SSA\textsuperscript{2}lign with specific hyper-parameter settings, supported by additional results of hyper-parameter sensitivity analysis to show the robustness of SSA\textsuperscript{2}lign. Subsequently, we present details of the cross-domain action recognition benchmarks for evaluating SSA\textsuperscript{2}lign, including Daily-DA and Sports-DA; lastly, we compare in detail our SSA\textsuperscript{2}lign with related but different FSDA and UDA/VUDA methods to highlight our novelty.}

\subsection*{Implementation Details}
\label{section:supp:implementation}

\begin{figure*}[!ht]
	\begin{center}
		\includegraphics[width=1.\linewidth]{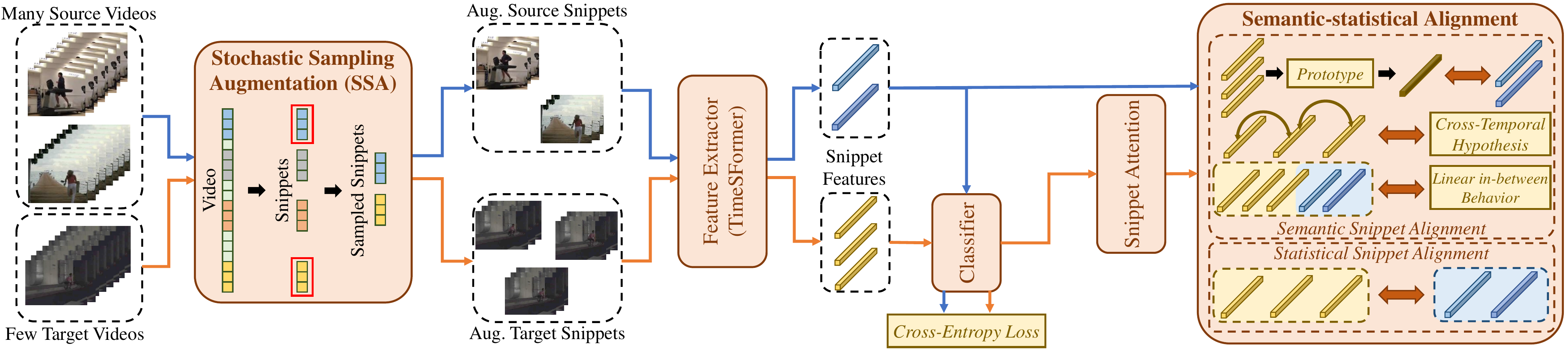}
	\end{center}
	\vspace{-6pt}
	\caption{Pipeline of SSA\textsuperscript{2}lign. Source and target snippets are first obtained through the Stochastic Sampling Augmentation, whose features are obtained through the shared feature extractor. SSA\textsuperscript{2}lign then aligns the source and target domains at the snippet level with the Semantic-statistical Alignment, while weighing the impact of different target snippets through snippet attention, whose weight is built based on the output prediction of target snippets, obtained from a shared classifier with source snippets. The blue and orange lines imply the data flow for source and target videos respectively. \textit{Best viewed in color.}}
	\label{figure:supp-1-struct}
\end{figure*}

\begin{figure*}[!t]
	\begin{center}
		\includegraphics[width=1.\linewidth]{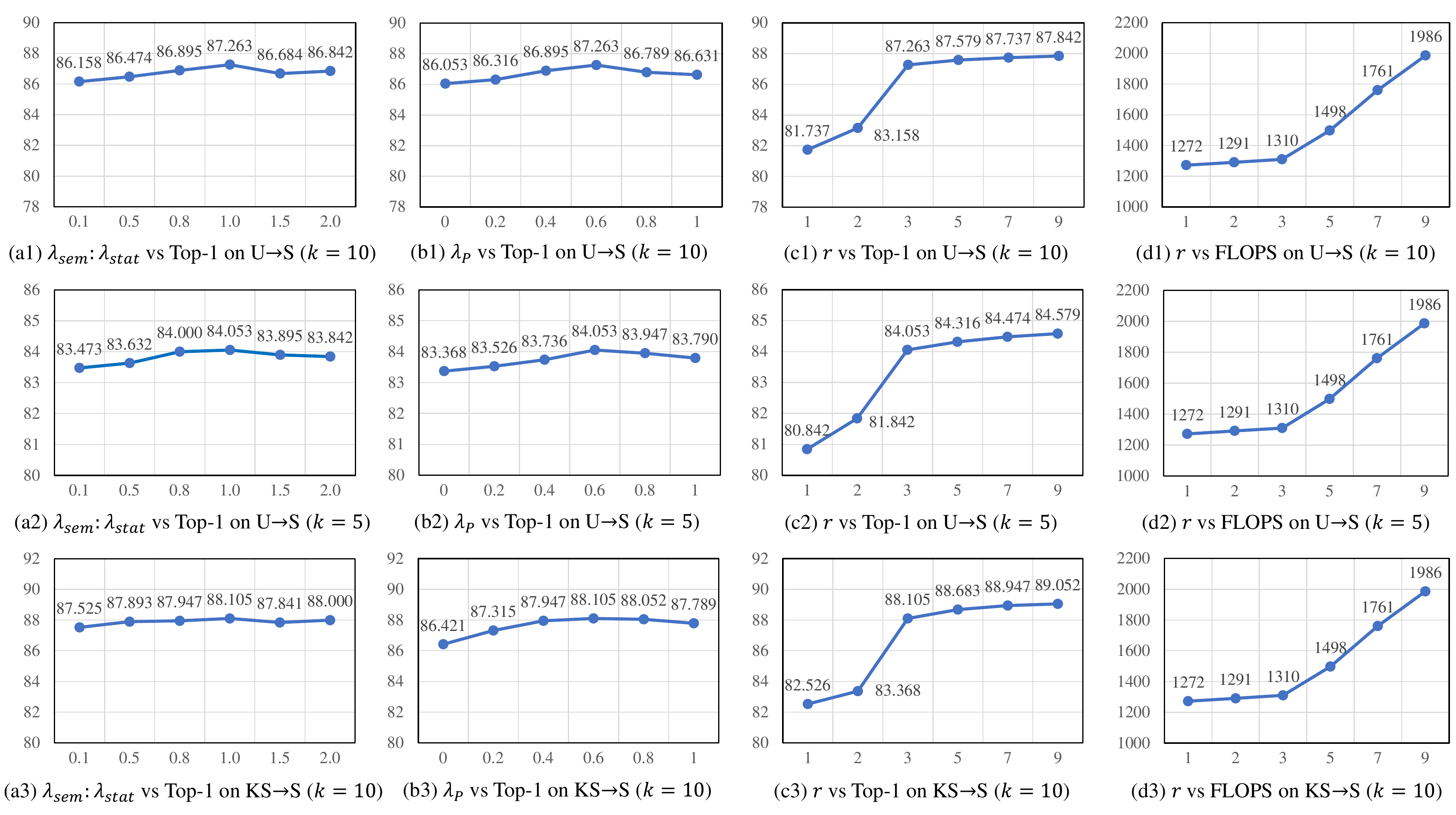}
	\end{center}
	\vspace{-12pt}
	\caption{Hyper-parameter sensitivity on the U$\to$S task with $k=10$ (top), $k=5$ (mid), and the KS$\to$S task with $k=10$ (bot).}
	\label{figure:supp-2-hyperparameter-more}
\end{figure*}

\noindent \textbf{Brief Review of SSA\textsuperscript{2}lign.}
In this work, we propose the \textbf{S}nippet-attentive \textbf{S}emantic-statistical \textbf{Align}ment with \textbf{S}tochastic \textbf{S}ampling \textbf{A}ugmentation (\textbf{SSA\textsuperscript{2}lign}) to address \textit{Few-Shot Video Domain Adaptation} (FSVDA) by augmenting the source and target domains and performing domain alignment at the snippet level. SSA\textsuperscript{2}lign firstly augments the source and target domain data by a simple yet effective stochastic sampling process that makes full use of the abundance of snippet information and then performs semantic alignment from three perspectives: alignment based on semantic information within each snippet, cross-snippets of each video, and across snippet-level data distribution. To further improve the stability of snippet-level alignment, a statistical alignment strategy is additionally adopted, while snippet attention is proposed to weigh the impact of different target snippets on the domain alignment dynamically. In this section, we present the detailed implementation of SSA\textsuperscript{2}lign, whose pipeline is demonstrated in Fig.~\ref{figure:supp-1-struct}.

\noindent \textbf{TimeSFormer as Feature Extractor.}
To obtain features from snippets during training and videos during testing, we instantiate the Transformer-based TimeSFormer~\cite{bertasius2021space} as the feature extractor thanks to its capability in obtaining features that include both spatial and temporal information. TimeSFormer extracts spatial and temporal features with separate space-time attention blocks based on self-attention~\cite{vaswani2017attention} and obtains very competitive results on various action recognition benchmarks~\cite{bertasius2021space}. While other Transformer-based video models, such as Swin~\cite{liu2022video} and ViViT~\cite{arnab2021vivit}, also achieve competitive performances on action recognition, TimeSFormer possesses the least amount of parameters, requiring only $60\%$ parameters of Swin and only $40\%$ parameters of ViViT. The final classifier is implemented as a single fully connected layer. Both the feature extractor and the subsequent classifier are shared across source and target data.

\noindent \textbf{Training Details and Hyper-parameters.}
For training, we initialize the TimeSFormer feature extractor from pre-trained weights obtained by pre-training on Kinetics-400~\cite{kay2017kinetics}. For more efficient training, we freeze the first 8 blocks of TimeSFormer, leaving the last 4 blocks to be fully trainable, with the learning rate set at 0.005. ll new layers are trained from scratch, with their learning rates set to be 10 times that of the pretrained-loaded trainable layers (blocks). For the tasks constructed from the Daily-DA dataset~\cite{xu2023multi}, we train a total of 30 epochs, while we train a total of 50 epochs for tasks constructed from the Sports-DA dataset~\cite{xu2023multi}. The stochastic gradient descent (SGD) algorithm~\cite{bottou2010large} is used for optimization, with the weight decay set to 0.0001 and the momentum set to 0.9. During the training phase of SSA\textsuperscript{2}lign, the batch size is set to 24 input snippets per GPU, with 12 source snippets from 12 source videos and 12 target snippets from 4 target videos ($r=3$ by default). For a fair comparison, the batch size is set to 24 input videos per GPU when training all comparing methods. All experiments are implemented with the PyTorch~\cite{paszke2019pytorch} library and conducted on 2 NVIDIA A6000 GPUs. We set the length of snippets and the number of snippets per target video via SSA empirically as $m=8, r=3$. Hyper-parameters $\lambda_{sem}=1.0$, $\lambda_{stat}=1.0$ and $\lambda_{P}=0.6$ are empirically set and are fixed. As shown in Section 4.3 and Fig. 2 of the paper, the performance of SSA\textsuperscript{2}lign is robust to hyper-parameters $\lambda_{sem}$, $\lambda_{stat}$ and $\lambda_{P}$ as well as $r$ when $r\geqslant3$, with minimal variations and maintains the best results with high computation efficiency with all the default hyper-parameter settings. To further illustrate the robustness of SSA\textsuperscript{2}lign towards the sensitivity of $\lambda_{sem}$ and $\lambda_{stat}$ which control the strength of the semantic and statistical snippet alignment losses, $\lambda_{P}$ which relates to the update of target prototypes and $r$ the number of snippets per target video, we present the additional results of hyper-parameter sensitivity analysis under different experimental settings. Specifically, we present the results of the U$\to$S task with $k=10$, $k=5$ (the same as presented in Fig. 2 of the paper), and the results of the KS$\to$S task with $k=10$, as shown in Fig. \ref{figure:supp-2-hyperparameter-more} of this appendix.

The additional results further justify that SSA\textsuperscript{2}lign is robust to hyper-parameters $\lambda_{sem}$, $\lambda_{stat}$ and $\lambda_{P}$ as well as $r$ when $r\geqslant3$ under all examined experimental settings, while achieving the best results with high computation efficiency with the default hyper-parameter settings.

\subsection*{Cross-domain Action Recognition Benchmarks}
\label{section:supp:benchmarks}

\begin{table*}[!ht]
	\center
	\resizebox{.65\linewidth}{!}{
		\begin{tabular}{c|c|c}
			\hline
			\hline
			Statistics & Daily-DA & Sports-DA  \Tstrut\Bstrut\\
			\hline
			Video Classes \# & 8 & 23 \Tstrut\\
			Training Video \# & A:2,776 / H:560 / M:4,000 / KD:8,959 & U:2,145 / S:14,754 / KS:19,104\\
			Testing Video \# & A:1,289 / H:240 / M:400 / KD:725 & U:851 / S:1,900 / KS:1,961 \Bstrut\\
			\hline
			\hline
		\end{tabular}
	}
	\smallskip
	\caption{Summary of cross-domain action recognition benchmarks statistics.}
	\label{table:supp-1-stat}
\end{table*}

\begin{table*}[!ht]
	\center
	\resizebox{.95\linewidth}{!}{
		\smallskip\begin{tabular}{c|c|c|c|c}
			\hline
			\hline
			Class ID & ARID Class & HMDB51 Class & Moments-in-Time Class & Kinetics-600 Class \Tstrutlarge\Bstrutlarge\\
			\hline
			0 & Drink & drink & drinking & drinking shots \Tstrut\\
			1 & Jump & jump & jumping & jumping bicycle, jumping into pool, jumping jacks\\
			2 & Pick & pick & picking & picking fruit\\
			3 & Pour & pour & pouring & pouring beer\\
			4 & Push & push & pushing & pushing car, pushing cart, pushing wheelbarrow, pushing wheelchair\\
			5 & Run & run & running & running on treadmill\\
			6 & Walk & walk & walking & walking the dog, walking through snow\\
			7 & Wave & wave & waving & waving hand \Bstrut\\
			\hline
			\hline
		\end{tabular}
	}
	\smallskip
	\caption{List of action classes for Daily-DA.}
	\label{table:supp-2-daily-classes}
\end{table*}

\begin{table*}[!ht]
	\center
	\resizebox{.7\linewidth}{!}{
		\begin{tabular}{c|c|c|c}
			\hline
			\hline
			Class ID & UCF101 Class & Sports-1M Class & Kinetics-600 Class \Tstrut\Bstrut\\
			\hline
			0 & Archery & archery & archery \Tstrut\\
			1 & Baseball Pitch & baseball & catching or throwing baseball, hitting baseball\\
			2 & Basketball Shooting & basketball & playing basketball, shooting basketball\\
			3 & Biking & bicycle & riding a bike\\
			4 & Bowling & bowling & bowling\\
			5 & Breaststroke & breaststroke & swimming breast stroke\\
			6 & Diving & diving & springboard diving\\
			7 & Fencing & fencing & fencing (sport)\\
			8 & Field Hockey Penalty & field hockey & playing field hockey\\
			9 & Floor Gymnastics & floor (gymnastics) & gymnastics tumbling\\
			10 & Golf Swing & golf & golf chipping, golf driving, golf putting\\
			11 & Horse Race & horse racing & riding or walking with horse\\
			12 & Kayaking & kayaking & canoeing or kayaking\\
			13 & Rock Climbing Indoor & rock climbing & rock climbing\\
			14 & Rope Climbing & rope climbing & climbing a rope\\
			15 & Skate Boarding & skateboarding & skateboarding\\
			16 & Skiing & skiing & skiing crosscountry, skiing mono\\
			17 & Sumo Wrestling & sumo & wrestling\\
			18 & Surfing & surfing & surfing water\\
			19 & Tai Chi & t'ai chi ch'uan & tai chi\\
			20 & Tennis Swing & tennis & playing tennis\\
			21 & Trampoline Jumping & trampolining & bouncing on trampoline\\
			22 & Volleyball Spiking & volleyball & playing volleyball \Bstrut\\
			\hline
			\hline
		\end{tabular}
	}
	\smallskip
	\caption{List of action classes for Sports-DA.}
	\label{table:supp-3-sports-classes}
\end{table*}

In this paper, to evaluate our proposed SSA\textsuperscript{2}lign, we utilized two cross-domain action recognition benchmarks: the Daily-DA and Sports-DA~\cite{xu2023multi}. In this section, we provide more details on each benchmark.

\begin{table*}[!ht]
	\center
	\resizebox{.95\linewidth}{!}{
		\begin{tabular}{m{.14\textwidth}|m{.1\textwidth}|m{.35\textwidth}|m{.45\textwidth}}
			\hline
			\hline
			Method & Publication & Task & Techniques \Tstrut\Bstrut\\
			\hline
			d-SNE~\cite{xu2019dsne} & CVPR-19 & Few-Shot Domain Adaptation (FSDA): source image data available with labels, \textbf{a few} (very limited) target image data available with labels, image-based.\Tstrut\Bstrut& (a) d-SNE learns a latent domain-agnostic space through SNE~\cite{hinton2002stochastic} with large-margin nearest neighborhood~\cite{domeniconi2005large}; (b) d-SNE conducts FSDA in a min-max formulation with a modified-Hausdorff distance; (c) d-SNE creates sibling target samples with spatial augmentations, and trains feature extractor with the Mean-Teacher technique~\cite{tarvainen2017mean}.\Tstrut\Bstrut\\
			\hline
			PASTN~\cite{gao2020pairwise-tip} & TIP-20 & Few-Shot Video Domain Adaptation (FSVDA): source video data  available with labels, \textbf{a few} (very limited) target video data available with labels, video-based.\Tstrut\Bstrut& (a) PASTN obtains video features from a frame-based video model; (b) PASTN forms source-target video pairs to address insufficient target video data; (c) PASTN constructs pairwise adversarial networks performed across source-target video pairs optimized by a pairwise margin discrimination loss~\cite{wu2017sampling}.\Tstrut\Bstrut\\
			\hline
			DM-ADA~\cite{xu2020adversarial} & AAAI-20 & Unsupervised Domain Adaptation (UDA): source image data available with labels, sufficient target images available without labels, image-based.\Tstrut\Bstrut& (a) DM-ADA augments the target domain with the source domain by domain mixup~\cite{zhang2018mixup}; (b) DM-ADA improves the feature extractor by leveraging soft domain labels; (c) DM-ADA jointly trains a domain discriminator which judges the samples' differences relative to the two domains with refined scores.\Tstrut\Bstrut\\
			\hline
			ACAN~\cite{xu2022aligning} & TNNLS-22 & Video Unsupervised Domain Adaptation (VUDA): source video data  available with labels, sufficient target videos available without labels, video-based.\Tstrut\Bstrut& (a) ACAN applies adversarial-based domain adaptation across spatio-temporal video features; (b) ACAN additionally aligns video correlation features in the form of long-range spatiotemporal dependencies~\cite{wang2018non}; (c) ACAN further aligns the joint distribution of correlation information of different domains by minimizing pixel correlation discrepancy (PCD).\Tstrut\Bstrut\\
			\hline
			\textbf{SSA\textsuperscript{2}lign (Ours)} & - & Few-Shot Video Domain Adaptation (FSVDA): source video data  available with labels, \textbf{a few} (very limited) target video data available with labels, video-based.\Tstrut\Bstrut& (a) SSA\textsuperscript{2}lign addresses FSVDA at the \textbf{snippet level} instead of the frame or video-levels; (b) SSA\textsuperscript{2}lign augments target domain data and the snippet-level alignments by a simple yet effective stochastic sampling of snippets; (c) SSA\textsuperscript{2}lign performs both semantic and statistical alignments attentively, with the semantic alignments achieved by alignment based on the semantic information within each snippet, cross-snippets of each video, and across snippet-level data distribution.\Tstrut\Bstrut\\
			\hline
			\hline
		\end{tabular}
	}
	\smallskip
	\caption{Detailed comparison of SSA\textsuperscript{2}lign with related but different FS(V)DA and (V)UDA methods.}
	\label{table:supp-4-compare-method}
\end{table*}

\subsubsection*{Daily-DA}
The Daily-DA dataset is a recently proposed cross-domain action recognition dataset for VUDA~\cite{xu2023multi}. It is more comprehensive and challenging compared to prior benchmarks such as UCF-Olympic~\cite{sultani2014human} and UCF-HMDB\textsubscript{\textit{full}}~\cite{chen2019temporal} which have resulted in saturated performance due to limited domains (only 2 domains in each dataset) and number of videos per domain. Daily-DA includes videos of daily actions from four domains and incorporates both normal videos and low-illumination videos. Specifically, Daily-DA is built from four datasets: the dark dataset ARID (A)~\cite{xu2021arid}, as well as HMDB51 (H), Moments-in-Time (M)~\cite{monfort2019moments}, and Kinetics-600 (KD)~\cite{carreira2018short}, which are video datasets widely used for action recognition benchmarking~\cite{pareek2021survey}. Compared with other action recognition datasets such as Moments-in-Time and Kinetics, ARID is comprised of videos shot under adverse illumination conditions, characterized by low brightness and low contrast. Statistically, the RGB mean and standard deviation values (std) of videos in ARID are much lower among datasets leveraged in Daily-DA~\cite{xu2022aligning}, which strongly suggests a larger domain shift between ARID and the other action recognition datasets. The Daily-DA includes a total of 16,295 training videos and 2,654 testing videos from 8 categories as listed in Table \ref{table:supp-1-stat}, with each category corresponding to one or more categories in the original datasets as demonstrated in Table \ref{table:supp-2-daily-classes}.

\subsubsection*{Sports-DA}
To further demonstrate the efficacy of our proposed SSA\textsuperscript{2}lign on large-scale cross-domain datasets, we further adopt the Sports-DA dataset as another cross-domain action recognition benchmark. Comparatively, Sports-DA contains almost double the amount of training and testing videos of Daily-DA. Specifically, it includes a total of 36,003 training videos and 4,721 testing videos from 23 categories of sports actions, collected from three large-scale datasets: UCF101 (U)~\cite{soomro2012ucf101}, Sports-1M (S)~\cite{karpathy2014large}, and Kinetics-600 (KS)~\cite{carreira2018short}, as shown in Table \ref{table:supp-1-stat}. Similar to Daily-DA, each action class corresponds to one or more categories in the original datasets as presented in Table \ref{table:supp-3-sports-classes}. With more than 40,000 training and testing videos, the Sports-DA benchmark is one of the largest cross-domain action recognition benchmarks introduced.

\subsection*{Detailed Comparison with Related FS(V)DA and (V)UDA Methods}
\label{section:supp:detail-compare}
In this paper, we proposed SSA\textsuperscript{2}lign to address the more realistic and challenging FSVDA task, which achieves state-of-the-art performances with outstanding improvements on both cross-domain action recognition benchmarks (average $13.1\%$ on Daily-DA tasks and average $4.2\%$ on Sports-DA tasks). To further highlight the novelty of SSA\textsuperscript{2}lign, we compare our proposed SSA\textsuperscript{2}lign with prior FSDA/FSVDA and UDA/VUDA methods. Specifically, we compare with d-SNE~\cite{xu2019dsne} proposed for FSDA, PASTN~\cite{gao2020pairwise-tip} designed for FSVDA, ACAN~\cite{xu2022aligning} introduced for ACAN, and DM-ADA~\cite{xu2020adversarial} which is an image-based UDA method that leverages MixUp~\cite{zhang2018mixup}. These methods are all compared from two perspectives: the tasks they tackle and the techniques leveraged, as displayed in Table~\ref{table:supp-4-compare-method}.

{\small
\bibliographystyle{ieee_fullname}
\bibliography{iccv23}
}

\end{document}